\documentclass[runningheads]{llncs}
\usepackage[T1]{fontenc}
\usepackage{graphicx}
\usepackage{booktabs}
\usepackage{threeparttable}
\usepackage{multirow}
\usepackage{algorithmicx,algorithm}
\usepackage{amsbsy}
\usepackage{amsfonts}
\usepackage{marvosym}
\usepackage{amsmath}
\usepackage{makecell}

\begin{document}
%
\title{OTFPF: Optimal Transport-Based Feature Pyramid Fusion Network for Brain Age Estimation with 3D Overlapped ConvNeXt}
\titlerunning{OTFPF}
\author{Yu Fu\inst{1\ast} \and  
		Yanyan Huang\inst{1\ast} \and  
		Yalin Wang\inst{2} \and      
		Shunjie Dong\inst{1} \and
		Le Xue\inst{1} \and \\
		Xunzhao Yin\inst{1} \and
		Qianqian Yang\inst{1} \and
		Yiyu Shi\inst{3} \and
		Cheng Zhuo\inst{1}\textsuperscript{(\Letter)}}  
%
%
\institute{Zhejiang University, Hangzhou, China\\
\email{\{yufu1994,yyhuang.top,sj\_dong,11918441,xzyin1,\\qianqianyang20,czhuo\}@zju.edu.cn}\and
Arizona State University, Tempe, USA\\
\email{ylwang@asu.edu}\and
University of Notre Dame, Notre Dame, USA\\
\email{yshi4@nd.edu}}

\maketitle              

\begin{abstract}

Chronological age of healthy brain is able to be predicted using deep neural networks from T1-weighted magnetic resonance images (T1 MRIs), and the predicted brain age could serve as an effective biomarker for detecting aging-related diseases or disorders. In this paper, we propose an end-to-end neural network architecture, referred to as optimal transport based feature pyramid fusion (OTFPF) network, for the brain age estimation with T1 MRIs. The OTFPF consists of three types of modules: Optimal Transport based Feature Pyramid Fusion (OTFPF) module, 3D overlapped ConvNeXt (3D OL-ConvNeXt) module and fusion module. These modules strengthen the OTFPF network's understanding of each brain's semi-multimodal and multi-level feature pyramid information, and significantly improve its estimation performances. Comparing with recent state-of-the-art models, the proposed OTFPF converges faster and performs better. The experiments with 11,728 MRIs aged 3-97 years show that OTFPF network could provide accurate brain age estimation, yielding mean absolute error (MAE) of 2.097, Pearson's correlation coefficient (PCC) of 0.993 and Spearman's rank correlation coefficient (SRCC) of 0.989, between the estimated and chronological ages. Widespread quantitative experiments and ablation experiments demonstrate the superiority and rationality of OTFPF network. The codes and implement details will be released on GitHub: \url{https://github.com/brain-age/OTFPF} after final decision.

\end{abstract}

\vspace{-28pt}
\section{Introduction}

Brain aging, accompanied by complex biological and neuroanatomical changes, is a continuous and lifelong process~\cite{bashyam2020mri}. The main application prospect of the brain age estimation models is that it can be trained to predict the brain age as close as possible to the actual chronological age in the healthy brain, and then be applied to abnormal brain image data to reflect the accelerated or decelerated brain aging~\cite{cheng2021brain,he2021multi}. Based on the noninvasive imaging technology, magnetic resonance images (MRIs) own rich brain morphological information and have been widely adopted as an informative biomarker in deep learning community for brain age estimation~\cite{bashyam2020mri,he2021multi,peng2021accurate,cheng2021brain,bellantuono2021predicting}. Aiming to learn the correspondence between T1 and age labels, brain age estimation algorithms usually serve as high-dimensional regression models that predict chronological ages from 3D T1 images. Existing methods mainly adopt different 2D, 3D convolutional neural networks (CNNs) or their corresponding hybrid variants. The key measurement to evaluate whether a model is suitable for brain age estimation is if it can achieve a smaller mean absolute error (MAE), a bigger Pearson correlation coefficient (PCC), and a bigger Spearman’s rank correlation coefficient (SRCC). Vishnu et al.~\cite{bashyam2020mri} applied the inception-ResNet framework and achieved the MAE of 3.702 in the LifespanCN dataset. Peng et al.~\cite{peng2021accurate} proposed simple fully convolutional network (SFCN) and won Predictive Analytics Competition 2019 (PAC-2019) with MAE=2.90. When the chronological ages of MRIs are 18-90 years, the predicated brain ages are generally around 4-5 years~\cite{cole2020multimodality}. A relatively larger training dataset and a smaller chronological age range in both train dataset and test dataset generally result in a better estimation with smaller MAE, as demonstrated by~\cite{bellantuono2021predicting,cole2020multimodality}. 


As pointed by~\cite{cheng2021brain}, the inherent properties of CNNs, such as ignore the feature maps of different scales, limits the model performance. He et al.~\cite{he2021global} proposed the combination of transformer and CNNs (global-local transformer network, GLTN) to fuse the global-context and local detailed information to alleviate the scale problem. Cheng et al.~\cite{cheng2021brain} proposed the two-stage-age-network (TSAN) that applied DenseNet as the backbone and achieved the MAE of 2.428, but the huge amount of parameters and the non-end-to-end framework might limit its applications. Also, many previous studies ignored the biological sex labels of MRIs, while the brains of males and females have differential structure information and aging patterns~\cite{kaczkurkin2019sex,vosberg2021sex}. Adding other modalities that derived from T1 MRIs (the so-called "semi-multimodal," see~\cite{peng2021accurate}), such as gray matter and white matter, may boost the estimation performance. However, to the best of our knowledge, there is no model that considers the multi-scale information, semi-multimodal information, and how to integrate them in an optimized way.

To address the limitations of recent brain age estimation models, we propose an end-to-end neural network architecture, referred to as OTFPF network. The OTFPF network directly takes in each 3D T1 MRI, its corresponding gray matter image, white matter image and sex label as inputs, fully exploiting each brain's semi-multimodal and multi-level feature pyramid information with optimal transport based feature pyramid fusion (OTFPF) module, 3D overlapped ConvNeXt (3D OL-ConvNeXt) module and fusion module. Comparing with recent competing models, OTFPF network converges faster and performs better. Widespread experimental results quantitatively demonstrate the superiority and rationality of OTFPF network on commonly used metrics (MAE, PCC and SRCC).

\vspace{-6pt}


\section{Method}
\begin{figure}[htbp]
\centering
\includegraphics[width=1\textwidth]{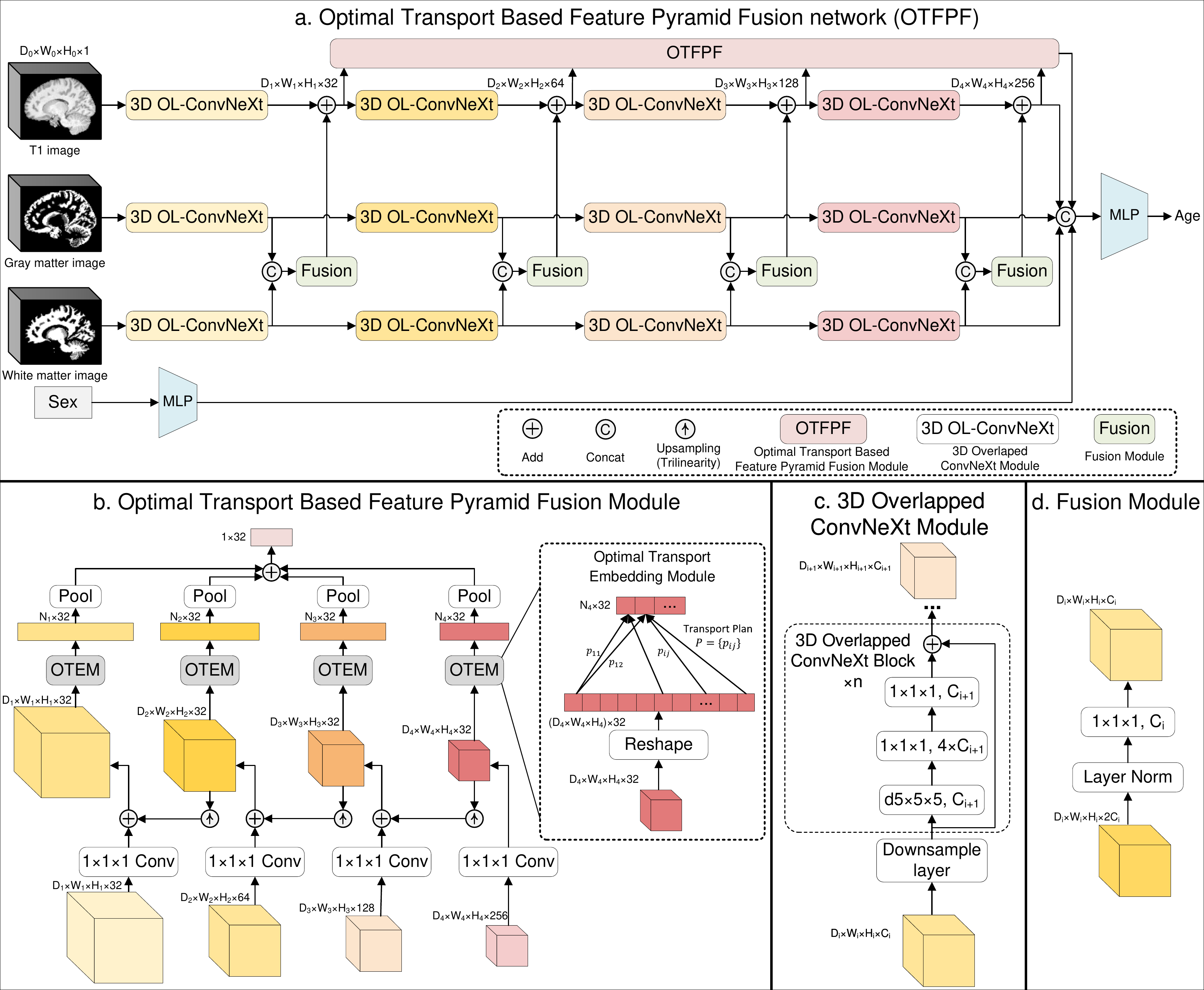}
\caption{(a) The overall architecture of OTFPF network, which consists of three types of modules: OTFPF module, 3D OL-ConvNeXt module and fusion module. (b) The detailed architecture of OTFPF module, which consists of two functional sub-modules: the feature pyramid fusion network (FPFN) and the optimal transport embedding module (OTEM). (c) 3D OL-ConvNeXt module. (d) Fusion module.} 
\label{Framework_BrainAge}
\vspace{-11pt}
\end{figure}

The overall architecture of OTFPF network is shown in Fig.~\ref{Framework_BrainAge}. It consists of three types of modules: OTFPF module, 3D OL-ConvNeXt module and fusion module. The OTFPF network has four pathways: a backbone pathway for T1 image, two branch pathways for gray matter image and white matter image, and one branch pathway for sex label. As shown in Fig.~\ref{Framework_BrainAge} (a), each of the first three pathways has four 3D OL-ConvNeXt modules. From left to right, the number of blocks of the four 3D OL-ConvNeXt modules is (1, 1, 3, 1). After each pass-through of the 3D OL-ConvNeXt module, the refined gray matter image features and white matter image features are concatenated and further sent to the fusion module for feature fusion. This process will be executed four times in a cascade manner to better explore multi-scale and semi-multimodal information. The semi-multimodal information of all fusion modules is added into the workflows in four stages of the backbone pathway to enhance its understanding of brain image. After the additions, these multi-scale and semi-multimodal features are fed into the OTPFF module to explore a suitable feature pyramid fusion strategy, and the feature aggregations are performed more effectively by leveraging the optimal transport embedding module (OTEM). Finally, the outputs of the four pathways are concatenated together and fed into a multilayer perceptron (MLP) to provide the final brain age estimation results. More detailed explanation of the OTFPF module, 3D OL-ConvNeXt module and fusion module can be found in the following sections.

\vspace{-16pt}
\subsection{Optimal Transport based Feature Pyramid Fusion Module}
Recent studies, such as~\cite{cheng2021brain,peng2021accurate,he2021multi}, only considered using feature maps after the final stage (i.e., the so called "higher-level features") of model to estimate the brain age. The features of initial or some middle stages are usually omitted, although they might be beneficial in the brain age estimation. After each stage, usually the spatial resolution is reduced and the number of features is increased~\cite{cheng2021brain,peng2021accurate,he2021multi}. It provides a cascade and pyramidal feature distribution patterns in different stages. To increase the prediction power of low-level, middle-level and high-level features, we propose an optimal transport based feature pyramid fusion (OTFPF) module, inspired by the recent success of feature pyramid network (FPN)~\cite{lin2017feature} and optimal transport architecture~\cite{peyre2019computational}. The proposed model may adaptively fuse these cascade and pyramidal features for more accurate and effective brain age estimations. The OTFPF module consists of two functional sub-modules: the feature pyramid fusion network (FPFN) and the optimal transport embedding module (OTEM). Details of the two sub-modules are described as follows. 

\vspace{-13pt}
\subsubsection{Feature Pyramid Fusion Network (FPFN).}
As shown in the Fig.~\ref{Framework_BrainAge} (a)-(b), the inputs of FPFN consists of four levels of features, which are denoted by: $f_{i}\in \mathbb{R}^{D_{i}\times W_{i}\times H_{i}\times C_{i}}, i\in \{1, 2, 3, 4\}$. Here the $(D_{i}\times W_{i}\times H_{i})$ and $C_{i}$ denote the three spatial dimensions and channels of features respectively, and $C_{1}=32, C_2=64, C_3=128, C_4=256$. Similar to FPN~\cite{lin2017feature}, the FPFN contains a bottom-up pathway and a top-down pathway. In the bottom-up pathway, point-wise convolutions are used for features at all levels, and the channel dimensions of all outputs are unified into $C=32$ for the sake of feature fusion in latter operations. The output features of bottom-up pathway are denoted by $f'_{i} \in \mathbb{R}^{D_{i}\times W_{i}\times H_{i}\times C_{1}}, i\in \{1, 2, 3, 4\}$. In the top-down pathway, the trilinear upsampling are applied for the $\{f'_2, f'_3, f'_4\}$ to get higher resolution features $\{f'_{21}, f'_{32}, f'_{43}\}$, which have the same spatial size with $\{f'_1, f'_2, f'_3\}$. Finally, these features of bottom-up pathway and top-down pathway are merged with add operations:$\{f''_1=f'_1+f'_{21}, f''_2=f'_2+f'_{32}, f''_3=f'_3+f'_{43}, f''_4=f'_4\}$.

\vspace{-13pt}
\subsubsection{Optimal Transport Embedding Module (OTEM).}
After the FPFN, we can get the fused pyramidal features: $\{f''_1, f''_2, f''_3, f''_4\}$. To adaptively fuse these features with different spatial dimensions, inspired by optimal transport kernel (OTK)~\cite{mialon2020trainable}, we propose the OTEM. The OTEM works in two steps: firstly, the OTEM reshapes the 3D tensors into 1D tensors, embeds these 1D tensors to a reproducing kernel Hilbert space (RKHS), and performs a weighted pooling with weights given by the transport plan; secondly, the embedded tensors of fixed length are obtained by using kernel approximation techniques~\cite{williams2000using}.


Refer to the Kantorovich relaxation of optimal transport with entropic regularization~\cite{peyre2019computational}. Let $\mathbf{a}$ and $\mathbf{b}$ be the weights of discrete measure $\alpha =\sum^n_{i=1}\mathbf{a}_i\delta_{x_i}$ and $\beta =\sum^m_{j=1}\mathbf{b}_j\delta_{y_j}$ with locations $x_i$ and $y_j$ respectively, and $\delta_z$ is the Dirac at position $z$. Denoting  $\mathbf{C}_{ij}=(c(x_i, y_j))_{ij}$ as the pairwise cost matrix evaluated on the support of $\alpha$ and $\beta$, and $c(x_i, y_j)$ is the ground cost for aligning the elements of $x_i$ and $y_j$. Then the entropic regularized Kantorovich relaxation of optimal transport from $x$ to $y$ is defined as:
\begin{equation}\small
    \setlength{\abovedisplayskip}{3pt}
    \mathbf{L}^{\varepsilon}_{\mathbf{C}}(\mathbf{a}, \mathbf{b})=\min\limits_{\mathbf{P}\in \mathbf{U}(\mathbf{a}, \mathbf{b})}\sum_{ij}\mathbf{C}_{ij}\mathbf{P}_{ij}-\varepsilon \mathbf{H}(\mathbf{P}) \label{OT}
    \setlength{\belowdisplayskip}{3pt}
\end{equation}
where $\mathbf{H}(\mathbf{P})=-\sum_{ij}\mathbf{P}_{ij}(\log(\mathbf{P}_{ij})-1)$ is the entropy of matrix $\mathbf{P}$, and $-\varepsilon\mathbf{H}(\mathbf{P})$ is the regularizing function with parameter $\varepsilon$, which controls the sparsity of $\mathbf{P}$. The $\mathbf{U}(\mathbf{a}, \mathbf{b})=\{\mathbf{P}\in \mathbb{R}^{n\times m}_+: \mathbf{P}\mathbf{1}_m=\mathbf{a}, \mathbf{P}^\mathrm{T}\mathbf{1}_n=\mathbf{b}\}$ is the space of admissible couplings between $\mathbf{a}$ and $\mathbf{b}$.This optimal transport problem is trying to solve $\mathbf{P}$, which is denoted as transport plan and carries the information on how to distribute the mass of $x$ in $y$ with minimal cost. This problem is typically solved by Sinkhorn's algorithm~\cite{cuturi2013sinkhorn}.

To be specific, we first reshape the output features $f''_{i}, i\in \{1, 2, 3, 4\}$ of the FPFN to 1D tensors: $f''_{ri}=(f''_{ri1}, ..., f''_{rim})\in \mathbb{R}^{D_i \cdot W_i \cdot H_i \times C_1}, m=D_i \cdot W_i \cdot H_i$, and then we define the optimal transport problem in this work as: embed features $f''_{ri}$ of varying large sizes to varying small fixed sizes representations $e_i=(e_{i1}, ..., e_{in_i}) \in \mathbb{R}^{n_i \times C_1}$, where $n_i$ is the number of elements in embedding. Inspired by the OTK~\cite{mialon2020trainable}, we first use positive definite kernel $\kappa$ to embed features to RKHS, and this operation is denoted as $\varphi:\mathbb{R}\rightarrow \mathcal{H}$. Then, we denote $m\times n$ matrix $\mathbf{P}(f''_{ri}, e_i)$ as the transport plan between $f''_{ri}$ and $e_i$, which is the unique solution of~\eqref{OT}, and the embedding is defined as:
\begin{equation}\small
\setlength{\abovedisplayskip}{3pt}
\begin{aligned}
    e_i=\Phi_{e_i}(f''_{ri}) &= \sqrt{n} \times \left(\sum^m_{j=1}\mathbf{P}(f''_{ri}, e_i)_{j1} \varphi (f''_{rij}),...,\sum^m_{j=1}\mathbf{P}(f''_{ri}, e_i)_{jn}\varphi (f''_{rij})\right)\\&=\sqrt{n}\times \mathbf{P}(f''_{ri}, e_i)^{\mathrm{T}}\varphi(f''_{ri})
\end{aligned}
\setlength{\belowdisplayskip}{3pt}
\end{equation}
where $\varphi(f''_{ri})=[\varphi (f''_{ri1},...,\varphi (f''_{rim}))]^{\mathrm{T}}$, and please note the output $e_i$ of expression (2) is not the final $e_i$ we used. 

In many cases, the embedding function $\varphi$ is infinite dimensional, which means we can't compute the embedding $\varphi(f''_{ri})$ explicitly, so the Nyström method is needed to implement approximately finite embedding $\psi(f''_{ri}) \in \mathbb{R}^{C_1}$. Specifically, we first projecting points from the RHKS $\mathcal{H}$ onto a linear sub-space $\mathcal{F}$, and the corresponding embedding admits an explicit form $\psi(f''_{ri})=\kappa(w,w)^{-1/2}\kappa(w,f''_{ri})$, where $\kappa(w, w)$ is the Gram matrix of kernel $\kappa$ computed on the set of anchor points $w$, and the anchor points can be chosen randomly or learned by back-propagation for supervised task~\cite{mairal2016end}. And thus, the finite embedding can be defined as:
\begin{equation}\small
\setlength{\abovedisplayskip}{3pt}
\begin{aligned}
    e_i=\Phi_{e_i}(f''_{ri}) &= \sqrt{n} \times \left(\sum^m_{j=1}\mathbf{P}(\psi(f''_{ri}), e_i)_{j1} \psi (f''_{rij}),...,\sum^m_{j=1}\mathbf{P}(\psi(f''_{ri}), e_i)_{jn}\psi (f''_{rij})\right)\\&=\sqrt{n}\times \mathbf{P}(\psi(f''_{ri}), e_i)^{\mathrm{T}}\psi(f''_{ri}) \in \mathbb{R}^{n_i \times C_1}
\end{aligned}
\setlength{\belowdisplayskip}{3pt}
\end{equation}
where the output $e_i$ of expression (3) is applied as our OTEM.


\vspace{-13pt}
\subsection{3D Overlapped ConvNeXt (3D OL-ConvNeXt) Module and Fusion Module}
Inspired by the the high-efficiency ConvNeXt framework~\cite{liu2022convnet}, we design the 3D OL-ConvNeXt module. As shown in Fig.~\ref{Framework_BrainAge} (c). each 3D OL-ConvNeXt module consists of ${n}$ 3D ConvNeXt blocks, which mainly downsamples the input feature maps in the three pathways of OTFPF. To our knowledge, the original ConvNeXt using patchify stem, which corresponds to non-overlapping convolution, and we found that may lose some important information considering the correlations of local regions of each brain. Therefore, we adopt an overlapped working mode for 3D ConvNeXt when downsampling, it significantly alleviates the information loss caused by dividing each global feature map into non-overlapping patches. As shown in Fig.~\ref{Framework_BrainAge} (d), the fusion module consists of a layer normalization operation and a point-wise convolution operation for a better feature fusion (e.g., remain the spatial dimensions of the features unchanged) of gray matter images and white matter images.

\vspace{-11pt}
\section{Experiments and Results}
\subsection{Setup}
\paragraph{Evaluation Datasets.} 
In this paper, we evaluate the proposed method on a healthy cohort: we collect the healthy brain T1-weighted MRIs from 11 public datasets (see Table~\ref{datasets}), with a total of 11,728 subjects aged 3-97 years. To train and evaluate different learning models, we randomly split the healthy cohort dataset into 3 subsets: the training set (80\%, 9382 MRIs ), validation set (10\%, 1173 MRIs), and test set (10\%, 1173 MRIs). The histogram matching and intensity normalization are performed for the harmonization of MRIs across datasets.

\begin{table}[t]\scriptsize
 \centering
 \caption{Demographic information of brain age estimation datasets used in this study.}
 \label{datasets}
 \setlength{\tabcolsep}{1.6mm}{
 \begin{threeparttable}
 \begin{tabular}{lcccc}  
  \toprule   
  Dataset & $N_{img}$ & Age Range & \makecell[c]{Age \\ (Mean$\pm$SD)} & \makecell[c]{Gender \\ (Male/Female)}\\
  \midrule   
  ABIDE I~\cite{di2014autism} & 571 & 6-56 & 17.10±7.73 & 472/99\\  
  ABIDE II~\cite{di2017enhancing} & 542 & 6-64 & 14.53±9.35 & 380/162 \\  
  ADNI~\cite{jack2008alzheimer} & 1673 & 56-96 & 75.75±6.78 & 829/844 \\
  CoRR~\cite{zuo2014open} & 1567 & 6-88 & 25.26±15.37 & 767/800 \\
  DLBS~\cite{park2012neural} & 315 & 21-89 & 54.62±20.09 & 117/198 \\
  ICBM$^a$ & 337 & 18-80 & 35.33±15.15 & 172/165 \\
  IXI$^b$ & 497 & 20-86 & 49.57±16.28 & 228/269 \\
  NKI-RS~\cite{nooner2012nki} & 2396 & 6-85 & 36.93±22.37 & 981/1415 \\
  OASIS-3~\cite{lamontagne2019oasis} & 1945 & 42-97 & 70.23±9.21 & 847/1098 \\
  OpenfMRI$^c$ & 1391 & 3-84 & 23.88±12.17 & 614/777 \\
  SALD~\cite{wei2018structural} & 494 & 19-80 & 45.18±17.44 & 185/309 \\
  \midrule
  Overall & 11,728 & 3-97 & 44.19±25.79 & 5592/6136 \\
  \bottomrule  
 \end{tabular}
 \begin{tablenotes}[flushleft]
    \item $^a$ \url{www.loni.usc.edu/ICBM}.  
    \item $^b$ \url{https://brain-development.org/ixi-dataset/}.
    \item $^c$ \url{https://openfmri.org/}. 
 \end{tablenotes}
 \end{threeparttable}}
 \vspace{-15pt}
\end{table}


\vspace{-11pt}
\paragraph{Implementation Details.} 
We implement the proposed OTFPF network using PyTorch library, and carry out all experiments on a single NVIDIA GeForce RTX 3090 GPU. The AdamW optimizer~\cite{loshchilov2017decoupled} is used to train the OTFPF network with an initial learning rate of $5\times10^{-4}$ and a weight decay factor of $5\times10^{-4}$. The batch size is set to be 16, and all images are processed by using a standard preprocessing pipeline with FSL 6.0~\cite{woolrich2009bayesian,smith2004advances,jenkinson2012fsl}, including nonlinear registration to standard 2mm MNI space and brain extraction~\cite{smith2002fast,jenkinson2005bet2}. The gray matter images and white matter images are obtained by using subcortical structure segmentation pipeline nested in the FSL~\cite{patenaude2011bayesian}. After preprocessing, all images own spatial size of $91\times109\times91$mm with spatial resolution of 2 mm$^3$.

\vspace{-11pt}
\paragraph{Methods for comparison.} We compare the proposed OTFPF network with three recent state-of-the-art approaches for brain age estimation using T1 images, including: simple fully convolutional network (SFCN)~\cite{peng2021accurate}, global-local transformer network (GLTN)~\cite{he2021global} and two-stage-age-network (TSAN)~\cite{cheng2021brain}. For a fair comparison, all of these models are re-implemented according to the original papers.

\vspace{-6pt}

\subsection{Results}

\begin{figure}[htbp]
\centering
\includegraphics[width=11cm]{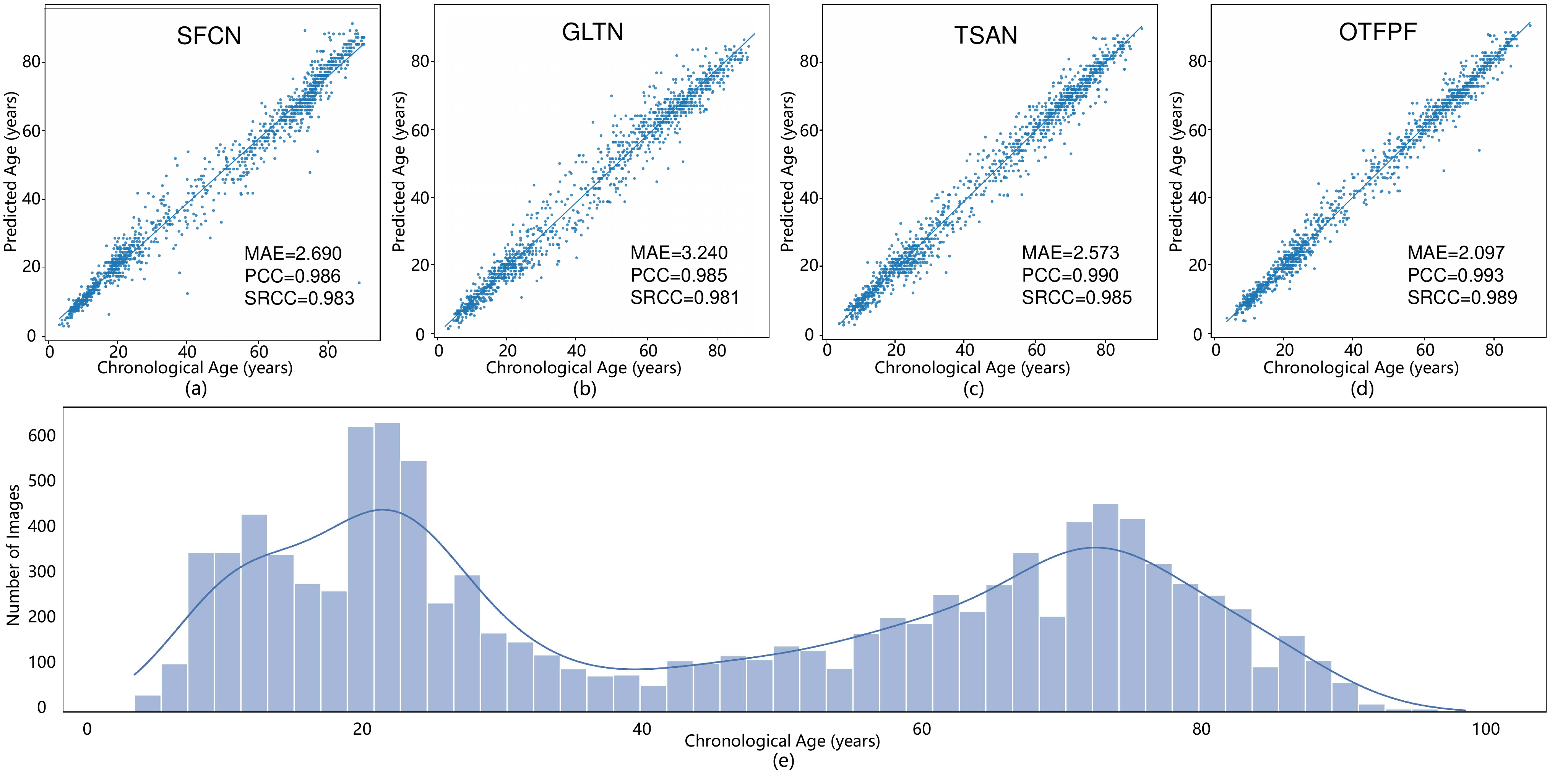}
\caption{(a)-(d) show the scatter diagrams of estimated brain ages by: (a) SFCN; (b) GLTN; (c) TSAN and (d) OTFPF. The blue line indicates the ideal estimation of $y = x$ (i.e., the estimated brain age equals to the chronological
age). (e) shows the chronological age histogram of all MRIs adopted in this study.} 
\label{scatter}
\end{figure}

Table~\ref{TestOnModels} details the quantitative brain age estimation results in the test set by OFFPF network and its competing models, respectively. We can note that the proposed OTFPF network achieves the lowest MAE of 2.097, the highest PCC of 0.993 and the highest SRCC of 0.989. These metrics demonstrate the OTFPF network achieves the smallest "age gap" between the estimated age and the actual chronological age. Fig.~\ref{scatter} (a)-(d) display the scatter diagrams of the estimated brain ages against the chronological brain ages based on SFCN, GLTN, TSAN and OTFPF network; Fig.~\ref{scatter} (e) shows the chronological age histogram of all MRIs adopted in this study. Relative to other three models, the OTFPF network can achieve a more stable estimation performance in subjects aged 40-60 (Fig.~\ref{scatter} (d)), which demonstrates the potential of OTFPF network in small samples (see Fig.~\ref{scatter} (e)).


\begin{table}[t]\scriptsize
 \centering
 \caption{Brain age estimation results by OTFPF and competing models.}
 \label{TestOnModels}
 \setlength{\tabcolsep}{5.8mm}{
 \begin{threeparttable}
 \begin{tabular}{lccc}  
\toprule 
Model & MAE & PCC & SRCC \\
\midrule 
SFCN & 2.690 & 0.986 & 0.983 \\
GLTN & 3.240 & 0.985 & 0.981 \\
TSAN & 2.573 & 0.990 & 0.985 \\
{\bf OTFPF} & {\bf 2.097} & {\bf 0.993} & {\bf 0.989} \\
\bottomrule
 \end{tabular}
 \begin{tablenotes}[flushleft]
    \item MAE: mean absolute error; PCC: Pearson correlation coefficient; SRCC: Spearman’s rank correlation coefficient.
 \end{tablenotes}
 \end{threeparttable}}
 \vspace{-15pt}
\end{table}


Widespread ablation experiments were performed. Results in Table~\ref{Ablation_Results} help us better understand the rationality and effectiveness of the modules and hyperparameters used in OTFPF network. OTFPF w/o (OTEM, FPFN and semi-multimodal information) performs the worst with the MAE of 2.397, PCC of 0.991 and SRCC of 0.986. Adding the semi-multimodal information (i.e., OTFPF w/o (OTEM and FPFN)) can decrease the MAE from 2.397 to 2.201, increase the PCC from 0.991 to 0.992 and increase the SRCC from 0.986 to 0.988. Only removing the OTEM (i.e., OTFPF w/o OTEM) rather than removing both OTEM and FPFN may degenerate the the MAE from 2.201 to 2.211, remain the PCC of 0.992 and SRCC of 0.988.

OTFPF w/o sex label degenerates the the MAE from 2.097 to 2.293, decreases the PCC from 0.993 to 0.990 and decreases the SRCC from 0.989 to 0.986, which highlights the importance of sex label for brain age estimation. Using classical number of blocks for each stage of ConvNeXt, such as (3, 3, 9, 3)~\cite{liu2022convnet} rather than our (1, 1, 3, 1), not only consumes more GPU resource, but degenerates the MAE from 2.097 to 2.270, decreases the PCC from 0.993 to 0.991 and decreases the SRCC from 0.989 to 0.987. Adopting regular 3D ConvNeXt (i.e., non-overlapping convolution) significantly degenerates the MAE from 2.097 to 2.718, decreases the PCC from 0.993 to 0.987 and decreases the SRCC from 0.989 to 0.982, which demonstrates that our designed 3D OL-ConvNeXt module can better explores the morphological information among brain regions. Overall, OTFPF outperforms all of its variants, demonstrating the potential and necessity of having these modules and hyperparameters in it. More detailed illustrations of these variants of OTFPF can be found in our supplementary materials.

\begin{table}[t]\scriptsize
 \centering
 \caption{Ablation experiments for both modules and hyperparameters.}
 \label{Ablation_Results}
 \setlength{\tabcolsep}{1.8mm}{
 \begin{threeparttable}
 \begin{tabular}{lccc}  
  \toprule 
  Model & MAE & PCC & SRCC \\  
  \midrule 
  OTFPF & 2.097 & 0.993 & 0.989 \\
  {\bf Ablation for modules}\\
  OTFPF w/o OTEM & 2.211 & 0.992 & 0.988 \\
  OTFPF w/o (OTEM and FPFN) & 2.201 & 0.992 & 0.988 \\ 
  OTFPF w/o (OTEM, FPFN and multi-pathway) & 2.397 & 0.991 & 0.986 \\ 
  {\bf Ablation for hyperparameters} &\\
  OTFPF w/o sex label & 2.293 & 0.990 & 0.986 \\
  OTFPF using classical stages of ConvNeXt & 2.270 & 0.991 & 0.987 \\
  OTFPF w/o 3D OL-ConvNeXt & 2.718 & 0.987 & 0.982\\
  \bottomrule 
 \end{tabular}
 \end{threeparttable}}
 \vspace{-15pt}
\end{table}

\vspace{-11pt}
\section{Conclusions}
\vspace{-8pt}
In this paper, we proposed the OTFPF network as an end-to-end neural network architecture for an accurate and effective brain age estimation with T1 MRIs. The OTFPF network can fully exploiting each brain's semi-multimodal and multi-level feature pyramid information to significantly improve the estimation performances. Comparing with recent competing models, OTFPF network converges faster and performs better. Numerical results demonstrate the rationality of each module in OTFPF network and show that the OTFPF network has good application prospects in future medical scenarios.


\bibliographystyle{splncs04}
\bibliography{paper910}

\end{document}